\documentclass[preprint,12pt]{elsarticle}

\usepackage{amssymb}
\usepackage{amsmath}
\usepackage{bm}

\usepackage{algorithm}
\usepackage[noend]{algpseudocode}
\algnewcommand{\LineComment}[1]{\State \(\triangleright\) #1}
\usepackage[colorlinks=true]{hyperref}

\journal{Robotics and Autonomous Systems}

\begin{document}

\begin{frontmatter}



\title{A Causal-based Approach to Explain, Predict and Prevent Failures in Robotic Tasks}


\author[label1]{Maximilian Diehl\corref{cor1}}
\ead{diehlm@chalmers.se}
\author[label1]{Karinne Ramirez-Amaro}
\cortext[cor1]{Corresponding Author}
\ead{karinne@chalmers.se}
\affiliation[label1]{organization={Department of Electrical Engineering, Chalmers University of Technology},
            city={Gothenburg},
            postcode={41296}, 
            country={Sweden}}

\begin{abstract}


Robots working in real environments need to adapt to unexpected changes to avoid failures. This is an open and complex challenge that requires robots to timely predict and identify the causes of failures to prevent them. In this paper, we present a causal method that will enable robots to predict when errors are likely to occur and prevent them from happening by executing a corrective action. First, we propose a causal-based method to detect the cause-effect relationships between task executions and their consequences by learning a causal Bayesian network (BN). The obtained model is transferred from simulated data to real scenarios to demonstrate the robustness and generalization of the obtained models. Based on the causal BN, the robot can predict if and why the executed action will succeed or not in its current state. Then, we introduce a novel method that finds the closest state alternatives through a contrastive Breadth-First-Search if the current action was predicted to fail. We evaluate our approach for the problem of stacking cubes in two cases; a) single stacks (stacking one cube) and; b) multiple stacks (stacking three cubes). In the single-stack case, our method was able to reduce the error rate by $97\%$. We also show that our approach can scale to capture multiple actions in one model, allowing to measure timely shifted action effects, such as the impact of an imprecise stack of the first cube on the stacking success of the third cube. For these complex situations, our model was able to prevent around $75\%$
of the stacking errors, even for the challenging multiple-stack scenario. Thus, demonstrating that our method is able to explain, predict, and prevent execution failures, which even scales to complex scenarios that require an understanding of how the action history impacts future actions. 
\end{abstract}

\begin{graphicalabstract}
\end{graphicalabstract}

\begin{highlights}
\item We propose a causal-based method that allows robots to understand possible causes for errors and predict how likely an action will succeed.
\item We then introduce a novel method that utilizes these prediction capabilities to find corrective actions which will allow the robot to prevent failures from happening.
\item Our algorithm proposes a solution to the complex challenge of timely shifted action effects. By detecting causal links over the history of several actions, the robot can effectively predict and prevent failures even if the root of a failure lies in a previous action.
\end{highlights}

\begin{keyword}
Causality in Robotics \sep Failure Prediction and Prevention \sep Explainable AI
\end{keyword}

\end{frontmatter}


\section{Introduction}
\label{sec:introduction}


%


Robots that act in human-centered environments have to handle the execution of various tasks, which requires them to adapt their plan execution flexibly to unexpected changes in the environment\cite{Kuestenmacher14}. Due to the complexity of these environments, we expect failures in various forms and for various reasons~\cite{Khalastchi18survey}, one example being execution failures. The ability to explain their own actions~\cite{bookOfWhy}, particularly when failures have occurred~\cite{Miller19Explanation, Hellstroem21}, is, therefore, an essential skill of such robots. However, diagnosis capabilities are not only crucial for detecting the causes~\cite{Mitrevski2020Workshop}, but could also be used to learn from failures and prevent them from happening~\cite{Mitrevski21FaultCorrection}. 

Generating explanations is conceptually based on causality methods~\cite{LewisCausalExplanation}, which are typically implemented through statistical techniques that learn a mapping between possible causes (preconditions) and the action-outcome (effect)~\cite{Mitrevski20IROS, Bauer20ICRA}. First, we need to investigate how robots can utilize prior experience to reason and consequently generate an explanation about what and why an action execution went wrong~\cite{DiehlRAL22}. For example, if a robot fails to execute the task of stacking a cube on top of another one, it should be able to explain that the execution failed because the upper cube was dropped too far to the left of the lower cube. In our previous work \cite{DiehlRAL22}, we proposed a causal-based method based on learning a causal Bayesian network to produce explanations when a failure was detected, meaning that the robot needed to fail. Upon failures, explanations were generated by contrastively comparing the variable parametrization associated with the failed activity with its closest parametrization that would have led to successful execution. Therefore, the next challenge is to use the acquired experience to predict failures in order to prevent them.


The causal relations obtained by the Bayesian networks can be used to predict how likely a particular parametrization of causes will produce failures. 
In this paper, we propose an extension of~\cite{DiehlRAL22} which makes use of the prediction capabilities of the learned BNs to prevent failures from happening. When the prediction of a failure has a high probability given the current state, our method finds an alternative execution state, which is expected to result in a successful action execution. This alternative state is found through BFS in a similar fashion as in~\cite{DiehlRAL22}, which allows the agent not only to prevent failures but, at the same time, to provide explanations for its corrective actions. 

\begin{figure}[t!]
\centering
  \includegraphics[width=\textwidth]{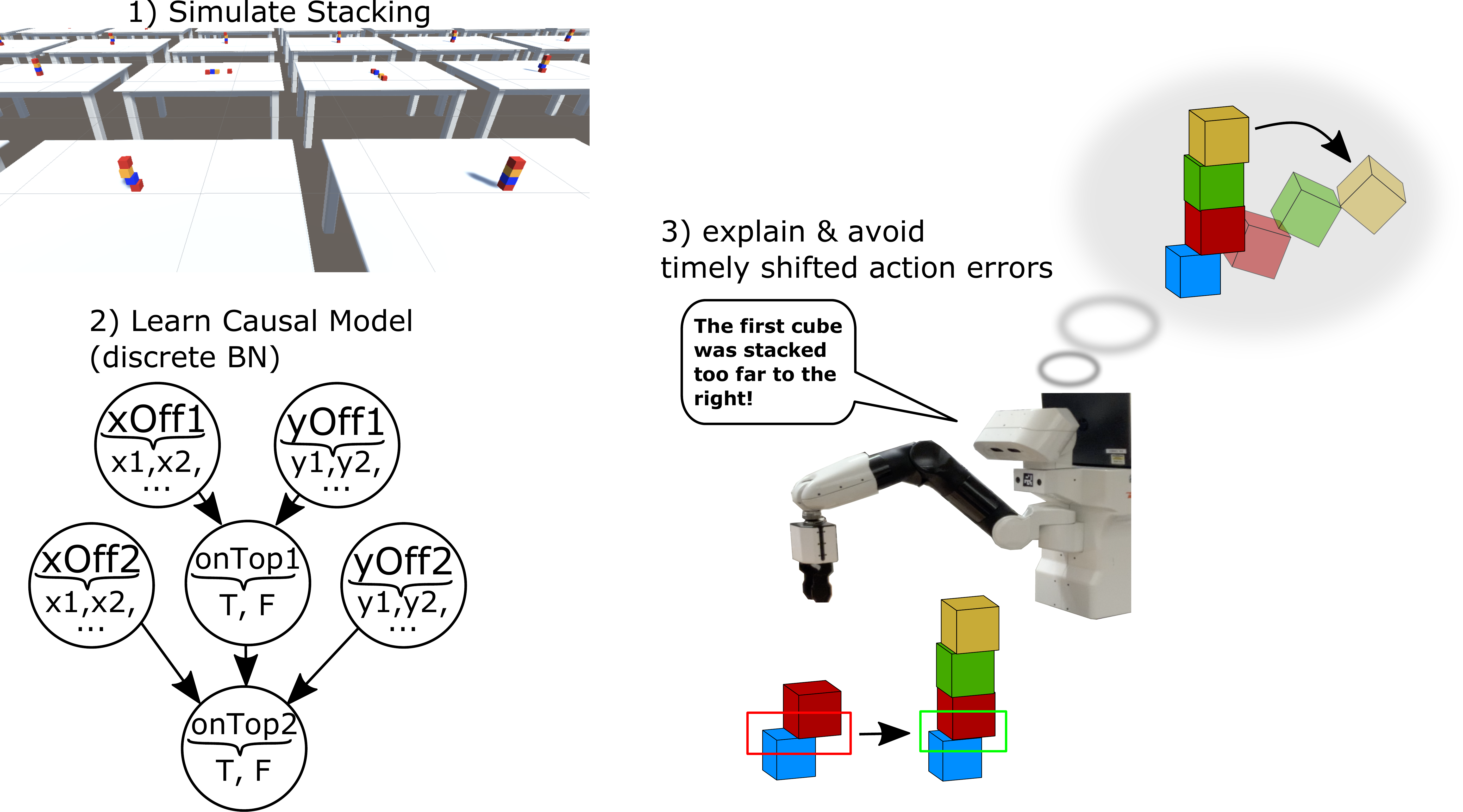}
  \caption{Depicts our method to allow robots to explain, predict, and prevent failures. First, a causal model is learned from simulations (steps 1,2). Then, this model is used to predict the success of an action given the current state and finds corrective actions in case the action is expected to fail (step 3), even in case of timely shifted action failures.}
\label{fig:overview}
\end{figure}

Predicting and preventing errors is particularly difficult if the effects of an action are not immediately flawed but become problematic in future actions~\cite{Altan2014ProbabilisticFI}. For example, the error was produced on the first action, but the consequence is only observed after the third action (in the future). We call these cases timely shifted action errors. In such cases, the models need to consider the history of the previous actions. Fig.~\ref{fig:overview} depicts the case of a robot building a tower of four cubes. The second (red) cube is not stacked entirely centered with respect to the bottom (blue) cube. Even if this particular stack can be considered successful on its own, it negatively impacts the overall stability of the tower, which might become a problem later after the second or third stack. This challenging problem is also addressed in this paper, and we show that our causal-based method scales to these complex cases by detecting causal links over the history of several actions, effectively predicting and preventing action failures. 


To summarize, our contributions are as follows: 
\begin{itemize}
    \item We propose a causal-based method that allows robots to understand possible causes for errors and predict how likely an action will succeed.
    \item We then introduce a novel method that utilizes these prediction capabilities to find corrective actions which will allow the robot to prevent failures from happening.
    \item Our algorithm proposes a solution to the complex challenge of timely shifted action effects. By detecting causal links over the history of several actions, the robot can effectively predict and prevent failures even if the root of a failure lies in a previous action.
\end{itemize}

\section{Related Work}
\label{sec:relatedWork}
\subsection{Causality in Robotics}
Even though the centrality of causality is increasingly recognized, it is still an underexplored topic in the robotics community~\cite{Brawer20IROS, Hellstroem21}. One of its significant strengths is the ability to discern task-relevant from irrelevant variables in data. This feature is taken advantage of, for example, in CREST~\cite{crest}, where causal interventions on environment variables are used to discover which variables affect an RL policy. Consequently, excluding irrelevant variables was found to positively affect generalizability and sim-to-real transfer of the policy. Another application was presented in~\cite{Bhat2016HumanoidIA}, where a set of task-agnostic learning rules was defined to learn causal relations in a physical task. In particular, through repeated interaction with its environment, a humanoid iCub robot learned a causal relationship between the weight of objects and its ability to increase the water level, while other variables, like color, were found to be irrelevant. Another paper~\cite{Uhde20RobotScientist} has the objective of learning causal relations between actions in household tasks. From human demonstrations in Virtual Reality, they discovered a causal link between opening a drawer and retrieving plates.
A causal approach to tool affordance learning was presented in~\cite{Brawer20IROS}. Their goal was to equip a robot with the ability to work with new tools more effectively through prior experience with different tools. Also~\cite{Song10} exploits the ability to learn the causal relevance of variables to discover dependencies between object, action, constraint features, and the task of grasp selection. Their main objective is to find the best grasp conditional on constraints and object features. 
The presented approaches and our method have in common that the framework of causality is used to a) detect causal links between certain preconditions in the environment and the success of an action~~\cite{crest, Bhat2016HumanoidIA, Uhde20RobotScientist} and b) make use of that knowledge to do something in an optimal fashion (e.g. picking the best tool~\cite{Brawer20IROS} or grasp~\cite{Song10}). However, those methods have not explored how this causal understanding can be used to explain, predict and prevent failures from happening. Furthermore none of these works discussed the problem of timely shifted action errors, which we are addressing in this paper. 

\subsection{Learning explainable models of cause-effect relations}
The planning community captures cause-effect relationships in the form of (probabilistic) planning operators~\cite{Diehl21}. Some works proposed the concept of task execution models, which combines symbolic preconditions and a function approximation for the success model~\cite{Mitrevski20IROS}, based on Gaussian Process models. They evaluate their method by learning how to grasp handles. 
The authors noted that a simulated environment could be incorporated for a faster and more extensive experience acquisition, as proposed in~\cite{Song10}, which is a technique we employ to learn our Bayesian Networks. Human virtual demonstrations have been used to construct planning operators to learn cause-effect relationships between actions and observed state-variable changes~\cite{Diehl21}. Bauer et al. learn probabilistic action effects of dropping objects into different containers~\cite{Bauer20ICRA}, with the goal of generalizing the probability predictions for a variety of objects, like bowls and bread boxes. They utilize an ontology to find out how closely related the objects are but do not consider object properties. Our approach has several advantages over the work mentioned in this subsection. While we share the general objective of learning generalizable prediction models that map preconditions to action effects, our proposed causal approach allows us, as opposed to~\cite{Bauer20ICRA}, to learn which object properties and environment preconditions are relevant for the investigated actions. This feature is also the basis for our method to explain, predict and prevent failures. Furthermore, our method scales to scenarios where the effect of a previous action has an impact on subsequent actions, as opposed to~\cite{Mitrevski20IROS}.

\subsection{Failure prevention}
In~\cite{Khalastchi18survey}, approaches for fault detection and diagnosis in Robotic systems are classified into data-driven, model-based, and knowledge-based. The ability to diagnose and correct robot action failures are discussed in~\cite{Mitrevski21FaultCorrection} for the problem of robot grasping. They propose a method that diagnoses potential causes (unmet preconditions) for a failure and, through sampling, finds a parametrization that maximizes predicted execution success under the known execution model. This method works well when the errors occur immediately after the action. However, if errors either occur in the future or build up cumulatively over several actions, we require a model that can capture the action history. In~\cite{Altan2014ProbabilisticFI} the authors propose a method that would also be able to detect the cause of failures, even if it is not related to the currently executed action. They utilize Hierarchical Hidden Markov Models (HHMMs) to represent and track failures over time, which allows them to generate a list of possible causes with different likelihoods when a failure is encountered during a plan execution. However, they have not utilized this information to prevent failure through corrective actions, as we propose. 

\subsection{Contrastive Explanations}
An important element of many explainable AI methods like Explainable AI Planning (XAIP)~\cite{XAIP}, is the concept of contrastive explanations. This concept draws parallels to the way that humans generate explanations~\cite{Miller19Explanation}. Typically the focus of work that falls under the umbrella of XAIP are questions like \textit{why the plan contains a particular action $a_1$ and not action $a_2$?}~\cite{XAIP, Seegebarth12} or they focus more on the actual communication of plan execution failures~\cite{Das21}. We utilize this concept in a different way and for a different purpose. We search for contrastive failure causes, which allow us to explain why failures might occur in the future and find corrective actions to prevent them from happening. 


\section{Our approach to explaining, predicting and preventing failures}
\label{sec:method}
We propose and present a multi-step approach to predicting and preventing failures, which consists of four main steps:
\begin{enumerate}
    \item We start by explaining the task of variable identification (sec.~\ref{sec:variabledef}). In this step, an action is represented in terms of a set of random variables that describe both possible preconditions as well as effects.
    \item Then, we learn a causal model using the identified variables from step 1 based on BN learning (sec.~\ref{sec:BN}). BN learning is typically divided into learning the causal connections between the variables (structure learning) and learning conditional probability distributions (parameter learning). 
    \item In section~\ref{sec:explain}, we elaborate on how we use the obtained causal model to predict when a failure is likely to occur and how we have previously used the model to explain the reasons for failures after they have occurred~\cite{DiehlRAL22}. 
    \item Finally, in section~\ref{sec:correction}, we further expand our method to address the problem of preventing failures. 
\end{enumerate}

\subsection{Variable definitions and assumptions}
\label{sec:variabledef}
Our method for explaining, predicting, and preventing failures is based on detecting causal relations between possible causes and effects of an action. Actions refer to concrete movements executed by the robot to act on its environment, e.g., reaching a cube or stacking a cube. For the purpose of learning the causal relations, we describe each action in terms of a set of random variables $\mathrm{\textbf{X}} = \{ X_1, X_2,..., X_n \}$. The choice of the number of ($n$) variables is, in principle, in the hands of the experiment designer. However, there are several aspects that need to be considered: We conceptually split $\mathrm{\textbf{X}}$ into a subset of treatment (cause) variables $C \subset \mathrm{\textbf{X}}$ and outcome (effect) variables $E \subset \mathrm{\textbf{X}}$. Then, the goal is to measure the effect of treatment variables on their outcome. In other words, $C$ and $E$ differ since we can decide and set values for variables in $C$, while outcome variables are not actively set but measured at the end or throughout the experiment. We measure the success of an action in terms of effect variables. For example, the action of cancer treatment is successful if the outcome variable that denotes whether a patient has cancer equals 0. Therefore, we necessarily need to define at least one effect variable.

We can collect data for learning causal models either from simulations or from the real world. A data sample $d$ consists of a particular parametrization of the previously defined set of variables $\mathrm{\textbf{X}}$, which we denote as $d = \{ X_1 = x_{1}, X_2 = x_{2}, ..., X_n = x_{n} \}$, where n denotes the number of variables.... We currently assume that $d$ contains a value for all variables in $\mathrm{\textbf{X}}$, but there are also Bayesian network learning methods that can deal with incomplete datasamples~\cite{DAGs}. We sample values for the causes $C$ randomly. Randomized controlled trials are referred to as the gold standard for causal inference~\cite{bookOfWhy} and allow us to avoid the danger of unmeasured confounders. Consequently, we can call the detected relations between the variables $\mathrm{\textbf{X}}$ causal and not merely correlations, which can be spurious. Then, one advantage is that the generated failure explanations are truly causal. The second advantage of causal models is that they can also answer interventional queries, whereas non-causal models can only answer observational queries. 

We define another set of variables $G \subset E$, which denote outcome variables via which the success of an action is specified and $X_{goal} = \{$ $d_{goal_1},$  $d_{goal_2},$ $...,d_{goal_h}\}$ a set that contains all possible variable parametrizations that denote a successful action execution. Each goal parametrization $d_{goal_l} \forall l \in \{1, 2, ..., h\}$, describes one possible variable assignment of a subset of the outcome variables $E$, that are possible successful action outcomes. For example, in the cancer treatment, there is only one single successful outcome, namely the patient is cancer-free, thus $h = 1$.   
Then, an action is successful \textit{iff} the parametrization $d_g = \{ X_{g_{1}} = x_{g_{1}}, X_{g_{2}} = x_{g_{2}},..., X_{g_{m}} = x_{g_{m}}\}$ of $G$, $d_g \in X_{goal}$, where $X_{g_{i}} \in G \quad \forall i = \{1,..., m\}$. Note that $G$ does not necessarily need to contain all variables $E$, but depends on the actual goal that one aims on measuring. Thus $m$ denotes the number of variables that are relevant for specifying the success of an action. It is out of scope of this paper to learn $X_{\text{goal}}$ and instead we assume it is provided. However, the robot has no a-priori knowledge about which variables in $\mathrm{\textbf{X}} = {X_1, X_2, ..., X_n}$ are in $C$ or $E$, nor how they are related. 


\subsection{Our proposed pipeline to learn causal models}
\label{sec:BN}
A Bayesian Network (BN) is defined as a \textit{directed acyclic graph} (DAG) $\mathcal{G} = (\mathrm{\textbf{V}}, A)$, where $\mathrm{\textbf{V}} = \{ X_1, X_2, ..., X_n \}$ represents a set of nodes which correspond to the random variables $\mathrm{\textbf{X}}$ that describe our action, and $A$ is the set of arcs~\cite{bnlearn} that denotes all relations between the variables. This dependency allows to factorize the \textit{joint probability distribution} of a BN into \textit{local probability distributions}, where each random variable $X_i$ only depends on its direct parents $\Pi_{X_i}$:
\begin{equation}\label{eq:bn} 
    P(X_1, X_2, ..., X_n) = \prod_{i=1}^{n}P(X_i|\Pi_{X_i})
\end{equation}
To learn a BN from data, we first learn the structure of the DAG, and then retrieve the local probability distributions, which is referred to as parameter learning.
Many structure learning algorithms cannot handle continuous variables as parents of categorical variables~\cite{ChenDisc17, bnlearn}. We, therefore, perform quantile discretization~\cite{bnR} on all continuous random variables in $\mathrm{\textbf{X}}$.
\subsubsection{Structure Learning}
To learn the causal relations $\mathcal{G} = (\mathrm{\textbf{V}}, A)$ between the variables we utilize the stable implementation of the PC~\cite{pc} algorithm. This algorithm is a \textit{constraint-based-algorithm}, and it uses statistical tests to determine conditional independence relations from the data~\cite{bnR}. PC is one of many structure learning algorithms~\cite{DAGs}, which could be used with equal eligibility for this step. In the following, we assume that the outcome of the structure learning step indeed reveals the correct graph $\mathcal{G}$. If $\mathcal{G}$ is not found to be correct or only partially directed, it might be required to collect more data samples or tune the number of discretization steps. 

\subsubsection{Parameter Learning}
The purpose of this step is to fit functions that reflect the \textit{local probability distributions}, of the factorization in formula \ref{eq:bn}. We utilize the bayes estimator for conditional probabilities to generate a conditional probability table:
\begin{equation}
    p(\theta | D) = \frac{p(D|\theta) p(\theta)}{\sum_{\theta} p(D|\theta) p(\theta)}.
\end{equation}
Our Bayes estimator uses a uniform prior that matches the Bayesian Dirichlet equivalent (bde) score~\cite{bnR}.

\subsubsection{Inference}
Given the network structure $\mathcal{G}$ and the conditional probability table, we can query the BN to retrieve information about the probability distribution of BN variables. This process is also called inference. We rely on logic sampling, which belongs to the family of approximate inference algorithms~\cite{bnR}.


\subsection{Our proposed method to explain failures}
\label{sec:explain}
In our previous work~\cite{DiehlRAL22}, we proposed a method to generate contrastive failure explanations, which uses the obtained causal Bayesian network to compute success predictions (summarized in algorithm~\ref{alg:explanation}). 

\setlength{\textfloatsep}{2pt}
\begin{algorithm}
  \caption{\label{alg:explanation}Get closest successful variable parametrization from causal model}
  \hspace*{\algorithmicindent} \textbf{Input:} failure variable parameterization $x_{\text{failure}}$, 
  structural equations $P(X_i|\Pi_{X_i})$, discretization intervals of all model variables $X_{\text{int}}$, success threshold $\epsilon$, goal parametrizations $X_{goal}$ \\
  \hspace*{\algorithmicindent} \textbf{Output:} solution variable parameterization $x_{\text{solution}_{\text{int}}}$, solution success probability prediction $p_{\text{solution}}$
  \begin{algorithmic}[1]
  \Procedure{GetClosestSuccIntervals}{$x_{\text{failure}}, P(X_i|\Pi_{X_i}), X_{\text{int}}, \epsilon, X_{goal}$}
      \State $x_{\text{current}_{\text{int}}} \leftarrow \textsc{getIntervalFromValues}(x_{\text{failure}}, X_{\text{int}})$
      \State $P \leftarrow \textsc{generateTransitionMatrix}(X_{\text{int}})$ 
      \State $q \leftarrow [x_{\text{current}_{\text{int}}]}$
      \State $v \leftarrow []$
      \While{$q \neq \emptyset$} 
       \State $node \leftarrow \textsc{Pop}(q)$ 
       \State $v \leftarrow \textsc{Append}(v, node)$
       \ForAll{$\text{transition } t \in P(node)$} 
       \State $child \leftarrow \textsc{Child}(P, node)$
       \If{$child \not\in q,v $}
       \State $p_{\text{solution}} = P(d_g \in X_{goal}|\Pi_{G}=child)$
       \If{$p_{\text{solution}} > \epsilon$}
       \State $x_{\text{solution}_{\text{int}}} \leftarrow child$
       \State $\textsc{return}(p_{\text{solution}}, x_{\text{solution}_{\text{int}}})$
       \Else
       \EndIf
       \State$ q \leftarrow \textsc{Append}(q, x_{\text{current}_{\text{int}}})$
       \EndIf
       \EndFor
       \EndWhile
   \EndProcedure
  \end{algorithmic}
\end{algorithm}

After retrieving the current interval from the continuous variable parametrization (L-1 Alg.~\ref{alg:explanation})), a transition matrix is generated (L-2 Alg.~\ref{alg:explanation})). This transition matrix will be used to provide all possible state transitions in the search tree. A state in the search tree is made up of a complete parametrization of all variables in $\mathrm{\textbf{X}}$. Let's consider the example of $\mathrm{\textbf{X}} = \{ X_1, X_2 \}$ with two intervals $x', x''$ each. Then, all possible valid transitions for $node = (X_1=x', X_2=x')$ would be $child_1 = (X_1=x', X_2=x'')$ or $child_2 = (X_1=x'', X_2=x')$.

In lines 5-15 (Alg. \ref{alg:explanation}), the closest variable parametrization that fulfills the goal criteria of $P(d_g \in X_{goal}|\Pi_{G}=child) > \epsilon$, is searched for through BFS. $\epsilon$ is the success threshold and can be set heuristically. 

\begin{figure}[ht!]
\centering
  \includegraphics[width=\textwidth]{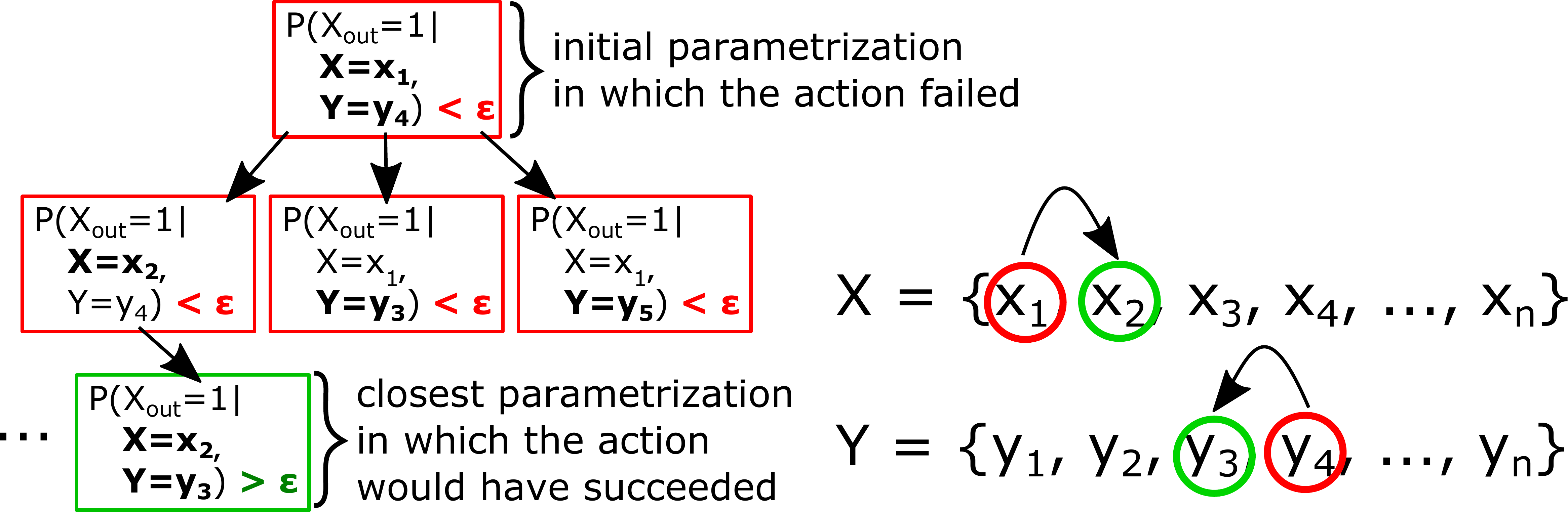}
  \caption{Exemplifies how contrastive explanations are generated from the BFS search tree. Figure taken from \cite{DiehlRAL22}.}
\label{fig:explainGen}
\end{figure}

The concept of our explanation generation is comparing the current variable intervals that lead to an execution failure  $x_{\text{current}_{\text{int}}}$ with the closest intervals would have been expected to lead to a successful task execution $x_{\text{solution}_{\text{int}}}$. This process is visualized in Fig.~\ref{fig:explainGen} exemplified on two variables $X$ and $Y$, which both have a causal effect on variable $X_{\text{out}}$. Given that $x_{\text{out}} = 1 \in X_{\text{goal}}$ would denote the succesfull action that is described through this causal model, the resulting explanation would be that the action has failed because $X=x_1$ instead of $X=x_2$ and $Y = y_4$ instead of $Y = y_3$.

\subsection{Our Proposed method to predict and prevent failures}
\label{sec:correction}
In~\cite{DiehlRAL22}, we have used Alg.~\ref{alg:explanation} to explain the reasons for a failure only after the failure has occurred. For this, our algorithm searches for the closest variable parametrization that would have likely led to successful execution. However, the causal model can also be used to predict the success probability of the current state prior to the actual execution. 

In this paper, we, therefore, propose an extension of our method presented in the new Alg.~\ref{alg:correction} to prevent failures from happening when an error has been predicted with a high probability. In particular, we first retrieve the discretization intervals for the current variable parametrization (L-1 Alg.~\ref{alg:correction})) and query the causal model to predict the success probability for the current state (L-2 Alg.~\ref{alg:correction})). In case the predicted probability is above a chosen threshold of $\epsilon$, we continue with the execution based on the current parameters (L-3 Alg.~\ref{alg:correction})). If, however, the probability is below the threshold (L-4 Alg.~\ref{alg:correction})), we retrieve the closest success parametrization through Alg.~\ref{alg:explanation} (L-7 Alg.~\ref{alg:correction})). Finally, we use the middle values of the corrected intervals as concrete parameters to retrieve a corrected action (L-8 Alg.~\ref{alg:correction})). Note, the algorithm only changes the parametrization for intervals that have been deemed responsible for the failure and keep the current parametrization for variables that are not problematic. The output parametrization $x_{success}$ can then be used to manipulate the environment to ensure the action will succeed, e.g., by moving the robot gripper into a different position.
\setlength{\textfloatsep}{2pt}
\begin{algorithm}
  \caption{\label{alg:correction}Predict and prevent failures}
  \hspace*{\algorithmicindent} \textbf{Input:} current variable parameterization $x_{\text{current}}$,
  structural equations $P(X_i|\Pi_{X_i})$, discretization intervals of all model variables $X_{\text{int}}$, success threshold $\epsilon$, goal parametrizations $X_{goal}$ \\
  \hspace*{\algorithmicindent} \textbf{Output:} Concrete success variable parametrization $x_{\text{success}}$
  \begin{algorithmic}[1]
  \Procedure{PreventFailures}{$x_{\text{current}}, P(X_i|\Pi_{X_i}), X_{\text{int}}, \epsilon$}
  \State $x_{\text{solution}_{\text{int}}} \leftarrow \textsc{getIntervalFromVal}(x_{\text{current}}, X_{\text{int}})$
  \State $p_{\text{solution}} = P(d_g \in X_{goal}|\Pi_{G}=child)$
  \State $x_{\text{success}} \leftarrow x_{\text{current}}$
  \If{$p_{\text{solution}} < \epsilon$}
    \State $x_{\text{failure}} = x_{\text{current}}$
     \State $p_{\text{solution}}, x_{\text{solution}_{\text{int}}} \leftarrow \newline
        \hspace*{4em} \textsc{GetClosestSuccIntervals}(x_{\text{failure}}, P(X_i|\Pi_{X_i}), X_{\text{int}}, \epsilon, X_{goal})$
      \State $x_{\text{success}} \leftarrow \textsc{MiddleValFromIntervals}(x_{\text{solution}_{\text{int}}}, x_{\text{current}}, X_{\text{int}})$
  \EndIf
  \State $\textsc{return}(x_{\text{current}})$
  \EndProcedure
 \end{algorithmic}
\end{algorithm}

\section{Experiments and Results}

We evaluate our method to predict and prevent execution failures for the problem of stacking cubes. We conducted two different experiments: \\
\textit{Experiment 1}: First, we evaluate our learned causal model in a simple task of stacking one cube, see Fig.~\ref{fig:environment}.a. From this task, we assess the correction abilities of the obtained causal model (see, Sec. \ref{sec:experiment1}).\\
\textit{Experiment 2}: Then, we assess our proposed method in a complex task of staking multiple cubes for building a tower of four cubes. With this experiment, we also investigate the case of performing the stacking action three times in a row (see Sec. \ref{sec:experiment2setup}).

For both experiments, we design an environment that contains two types of cubes: \textit{CubeDown} and $\textit{CubeUp}_i$, with $i$ being the stacking order of the upper cubes (e.g., $\textit{CubeUp}_1$ is the cube that is stacked first). \textit{CubeDown} is the bottom cube of a tower that, in our case, does not need to be rearranged or requires any movement prior to the stacking actions. All cubes have a fixed size of 5cm. To describe the stacking action we define four types of variables: $\texttt{xOff}_i$, $\texttt{yOff}_i$, $\texttt{dropOff}_i$, $\texttt{onTop}_i$, where $i$ denotes the corresponding cube. Fig.~\ref{fig:environment}.b and Fig.~\ref{fig:environment2}.b provides a detailed description of the variables for both experiments.


The data collection for training and evaluating the causal models is conducted in Unity3d, which employs the Nvidia PhysX engine for physics simulations. Inside Unity, we set up 400 parallel table environments to speed up the simulation process and data collection. At the beginning of every stacking experiment, the variable values for $\texttt{xOff}_i$, $\texttt{yOff}_i$, $\texttt{dropOff}_i$ are randomly sampled and the cube positions are initialized accordingly. Note, that the simulations are conducted without the existence of any robot and only involve cubes dropping from a predefined position. As a result, the simulations are less conservative than the real world. However, in~\cite{DiehlRAL22}, we have experimentally determined approximately 70\% congruence in terms of stacking success between simulated stacks and stacks that were performed by a real robot. 

\subsection{Experiment 1 Setup: Stack-1-Cube scenario}
\label{sec:experiment1}
\begin{figure}[ht!]
\centering
  \includegraphics[width=\textwidth]{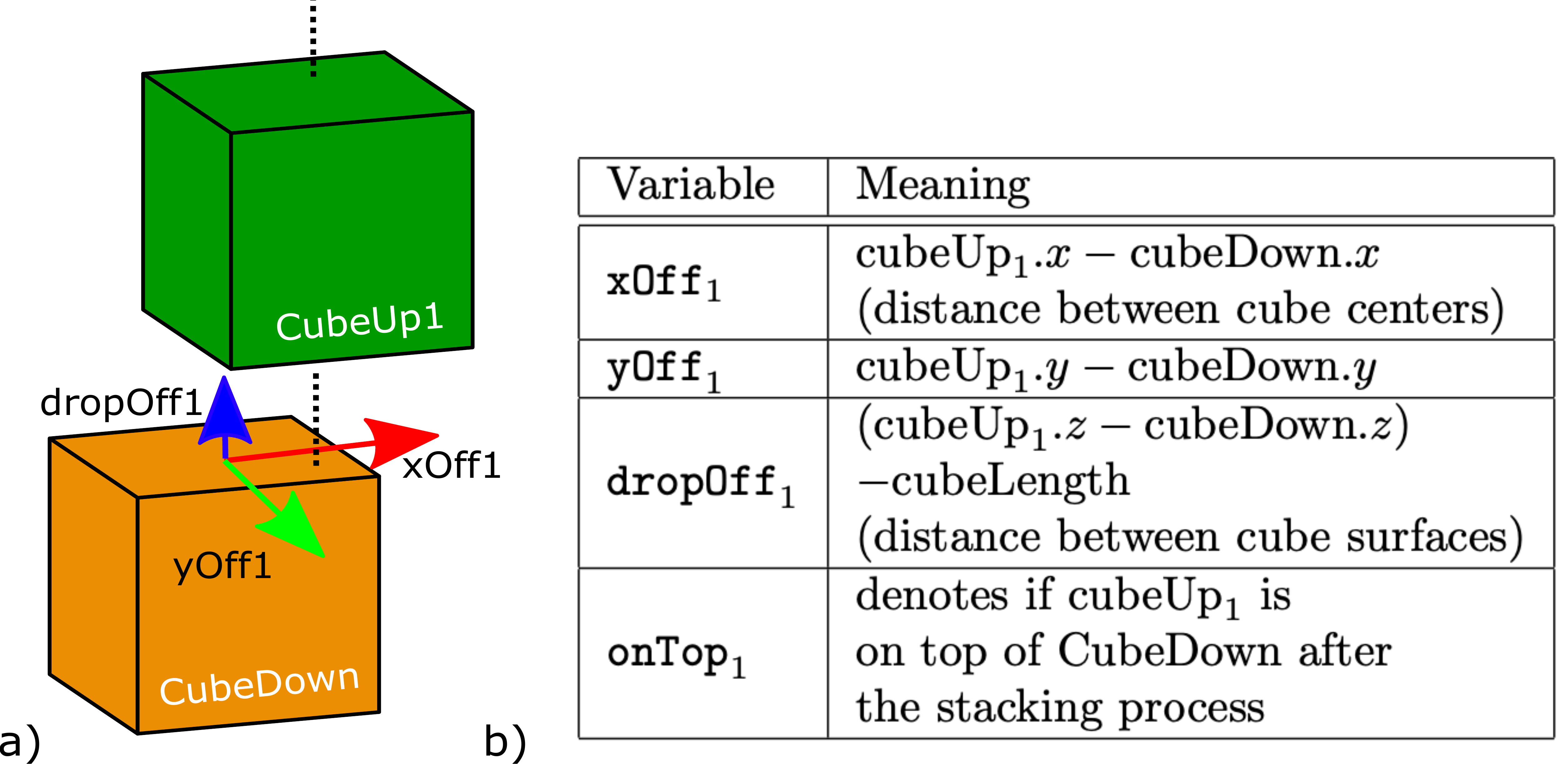}
  \caption{a) visualizes the variables that are used to describe that stacking action and b) defines their meaning.}
\label{fig:environment}
\end{figure}
In our first experiment, the goal is to stack only one single cube on top of the bottom cube (see Fig.~\ref{fig:environment}.a).
As the initial step, we deploy a training phase to collect data for learning the causal Bayesian network. For this purpose, we run 40,000 simulations of randomized single-stack actions. We define the set of variables that is used for the first experiment ($E1$) as $\mathrm{\textbf{X}}_{E1} = \{ \texttt{xOff}_i$, $\texttt{yOff}_i$, $\texttt{dropOff}_i$ $\texttt{onTop}_i \}$, with $i = \{ 1\}$. We sample randomly values for $\texttt{xOff}_i, \texttt{yOff}_i \sim \mathcal{U}_{[-0.03, 0.03]}$ (in meter), $\texttt{dropOff}_i \sim \mathcal{U}_{[0.005, 0.1]}$ (in meter). $\texttt{onTop}_i = \{ \text{True}, \text{False} \}$ is not sampled but automatically determined after the stacking process. From this training data, we learn the graphical representation of the variables and fit the conditional probability distributions. For each conditional value assignment, we then determine the closest variable parametrization that would lead to a successful execution based on the procedure which is elaborated in Sec.~\ref{sec:method}. This allows us to generate a 'lookup' table for the best possible corrections for each x-y-dropOffset parametrization that we could possibly encounter during a single stack action. 

The goal set $G = \{ \texttt{onTop}_1\}$ for this experiment, describes the stacking success for the cube that is stacked and consequently we denote the action as successful \textit{iff} $\texttt{onTop}_1=1$. We then evaluate the impact of having the obtained correction model in terms of cube-stacking success by comparing two cases (datasets): one without corrections from the model and one including the corrected stacking positions. Both test datasets use the same sample seed, different from the train-dataset seed. However, the variable distributions for training and testing are similar. Our hypothesis for this experiment is that deploying our method to adapt the stacking position prior to dropping the cube will significantly improve the stacking success. 

\subsection{Experiment 2 Setup: Stack-3-Cubes scenario}
\label{sec:experiment2setup}
\begin{figure}[ht!]
\centering
  \includegraphics[width=\textwidth]{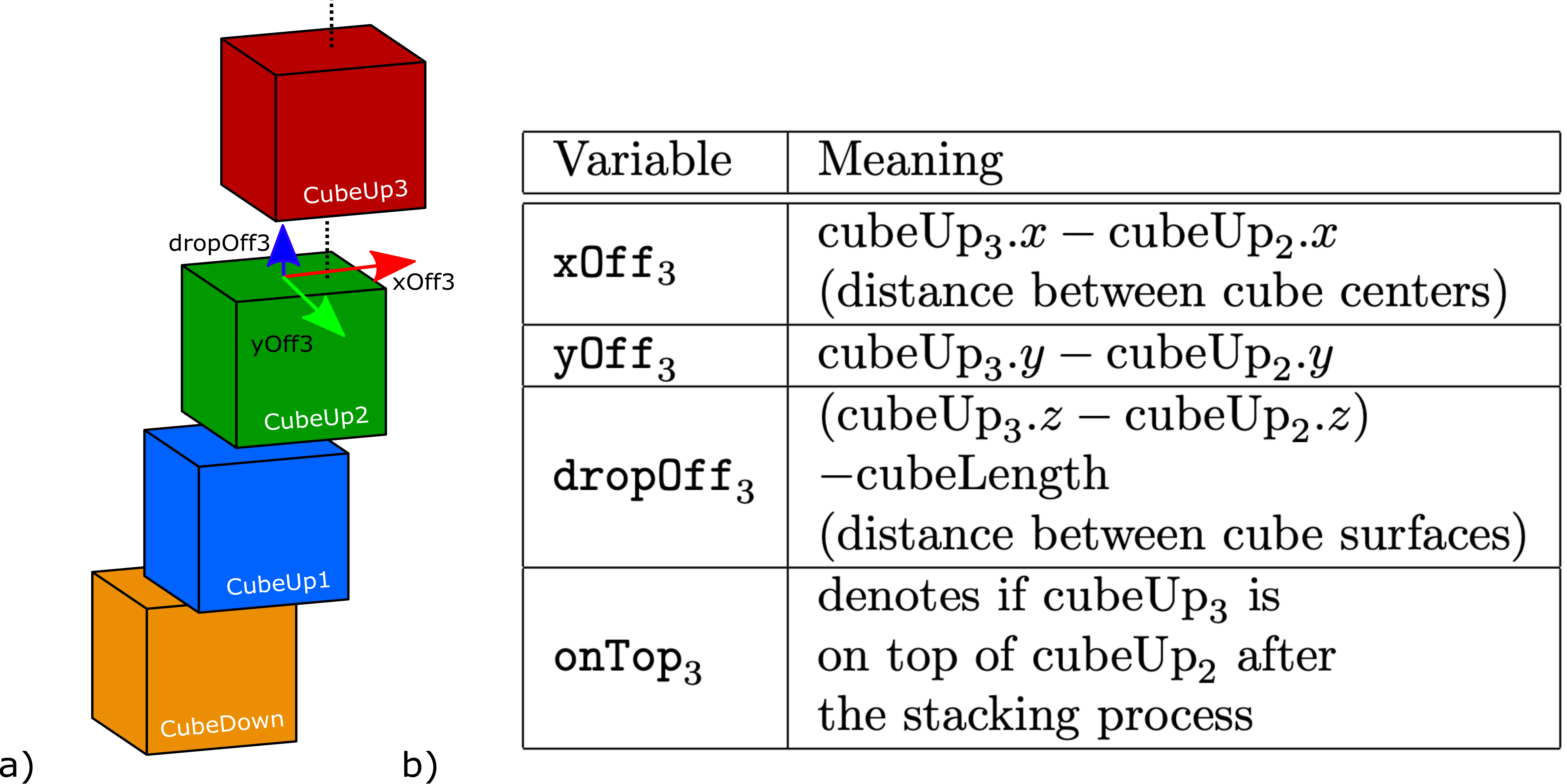}
  \caption{a) visualizes the variables that are used to describe multiple stacks and b) their meaning. Note, the offset variables are measured always with respect to the previous cube. }
\label{fig:environment2}
\end{figure}
Our second experiment ($E2$) considers the more complex scenario of stacking three cubes on top of a base cube (see Fig.~\ref{fig:environment2}.a). Unlike experiment 1, we need three upper cube variables: $\textit{CubeUp}_1$, $\textit{CubeUp}_2$ and $\textit{CubeUp}_3$ instead of a single upper cube. Now, the goal of the robot is to build a tower of cubes by stacking $\textit{CubeUp}_1$ on top of \textit{CubeDown}, $\textit{CubeUp}_2$ on top of $\textit{CubeUp}_1$ and, finally, $\textit{CubeUp}_3$ on top of $\textit{CubeUp}_2$. Our new set of variables is $\mathrm{\textbf{X}}_{E2} = $  $\{\texttt{xOff}_i$, $\texttt{yOff}_i$, $\texttt{dropOff}_i$, $\texttt{onTop}_i\}$, where $i = \{ 1, 2, 3\}$. In this case, we expect that the success of each stacking action becomes increasingly difficult, the higher the tower of cubes. 

To test this second experiment, we implement some slight adaptations to the simulation environment. For example, every 3 seconds, the next cube is dropped until the whole tower is complete. If a previous stack has failed, the whole experiment is considered a failure, and no more cubes are stacked on top, therefore, the experiment is terminated. Furthermore, the offset between two cubes is always calculated with respect to the previously stacked cube (e.g., between$\textit{CubeUp}_1$ and \textit{CubeDown} or $\textit{CubeUp}_2$ and $\textit{CubeUp}_1$), as exemplified in Fig.~\ref{fig:environment2}.b.

Similarly to the first experiment, we begin with a learning phase. Due to the requirement of the increased samples, we conducted a total of 800,000 experiments. In addition, unlike the uniformly distributed samples in experiment 1, we sample from a gaussian distribution for this experiment, to achieve a more equally distributed ratio between failures and successful stacks. Formally, $\texttt{xOff}_i, \texttt{yOff}_i \sim \mathcal{N}_{[0.0, 0.02]}$ (in meter), $\texttt{dropOff}_i \sim \mathcal{U}_{[0.001, 0.03]}$ (in meter) for $i = \{ 1, 2, 3 \}$. Again, $\texttt{onTop}_i = \{ \text{True}, \text{False} \}$ is not sampled but automatically determined after each stacking action. Also note that samples are limited to the ranges of $\texttt{xOff}_i, \texttt{yOff}_i = [-0.03, -0.03]$ and  $\texttt{dropOff}_i = [0.005, 0.1]$. This data is used to learn two different Bayesian networks: One only considers the first stacking action, thus representing a similar case as in experiment 1 ($i=1$), only trained on normally distributed data. For this first model, we take a subset of 40,000 samples. The second model, trained on all 800,000 samples, represents all three cubes ($i=1,2,3$) in one graphical representation.

We evaluate this experiment in two ways. First, we learn a BN only on the first stacking action, which we call $\texttt{1-Stack-Model}_g$ ($g$ denoting that the data was samples from Gaussian distributions), and we want to evaluate how useful this model is for failure prediction and prevention in later stacks. We don't reuse the model from experiment 1 ($\texttt{1-Stack-Model}_u$, where u=uniform distribution) since we expect some differences due to the adapted sampling distributions (E1:Uniform and E2: Gaussian distributions). Then, we learn a second model that captures all three stacks in one model, which we call the $\texttt{3-Stack-Model}_g$.
Consequently, we created three different test datasets, each consisting of 40,000 samples, following a similar distribution but other seed as in the training dataset. The first test dataset represents the case of no corrections (our baseline), followed by applying the smaller model on each of the cubes (no history case). Finally, we used the complete $\texttt{3-Stack-Model}_g$ to correct the data accordingly. 

The goal set $G = \{ \texttt{onTop}_3\}$ for this experiment indicates the stacking success of the third cube and we denote the action as successful \textit{iff} $\texttt{onTop}_3=1$. Note that, since the experiment was stopped, if one of the previous stacks has failed, a successful third stack will imply success in the other two stacks as well. We hypothesize that, while the $\texttt{1-Stack-Model}_g$ will improve the stacking success of the complete tower, the $\texttt{3-Stack-Model}_g$ will be even more helpful since it takes the entire history of all single stacking actions into account.

\subsection{Assessing the obtained causal models}
We first analyze the obtained causal models in terms of the graphical structure of the learned BN. Then, we explain the obtained conditional probabilities that were fitted around the experiment data. We validate the correctness of the model to make sure that the predictions are not based on a flawed understanding of cause-effect relations, which could result in wrong failure explanations and obstruct the failure prevention.  

\subsubsection{Obtained causal model for Experiment 1}
\begin{figure}[ht!]
\centering
  \includegraphics[width=0.4\textwidth]{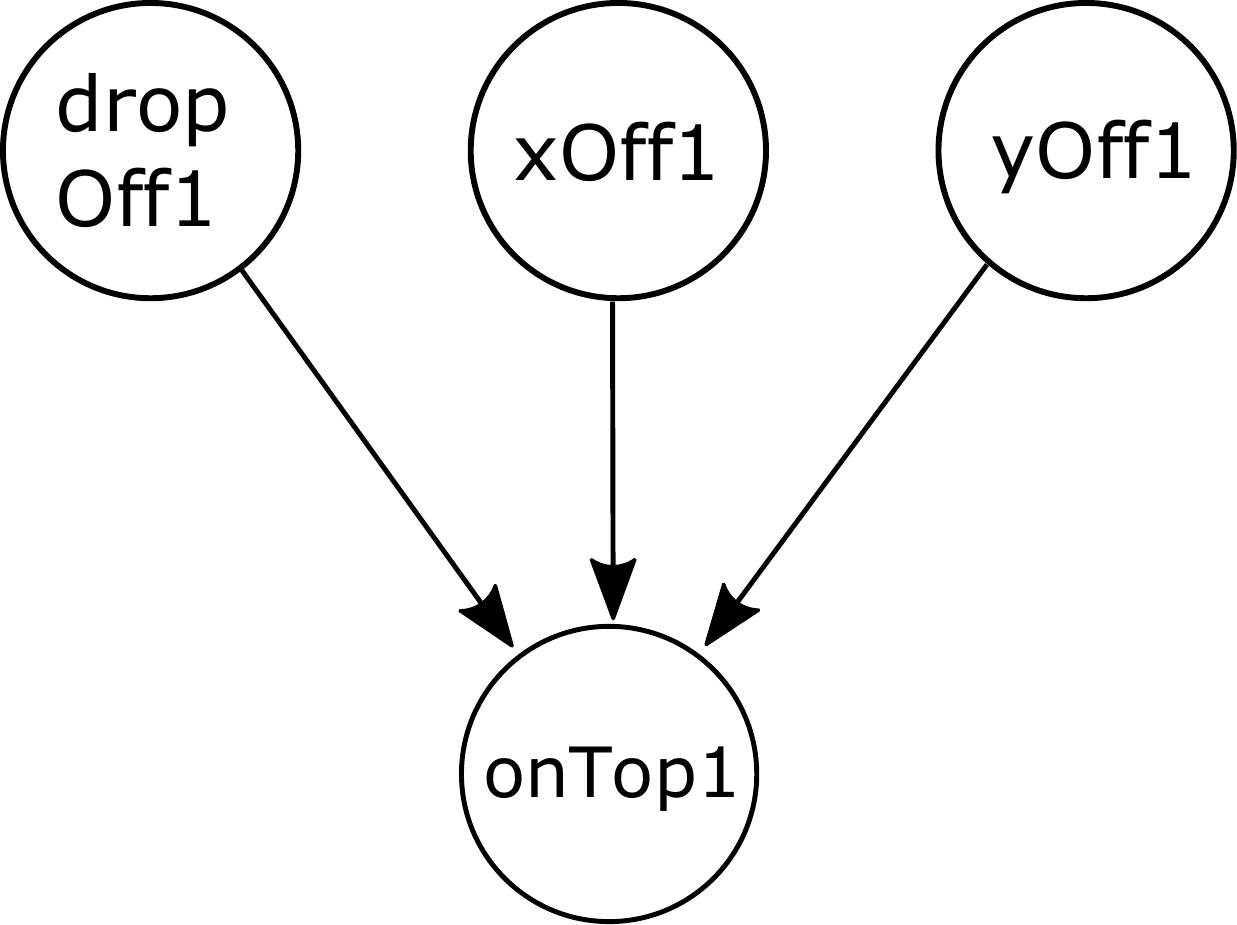}
  \caption{Obtained $\texttt{1-Stack-Model}_u$ Bayesian network structure for the Stack-1-Cube scenario.}
\label{fig:BNE1}
\end{figure}

\begin{figure}[ht!]
\centering
  \includegraphics[width=0.9\textwidth]{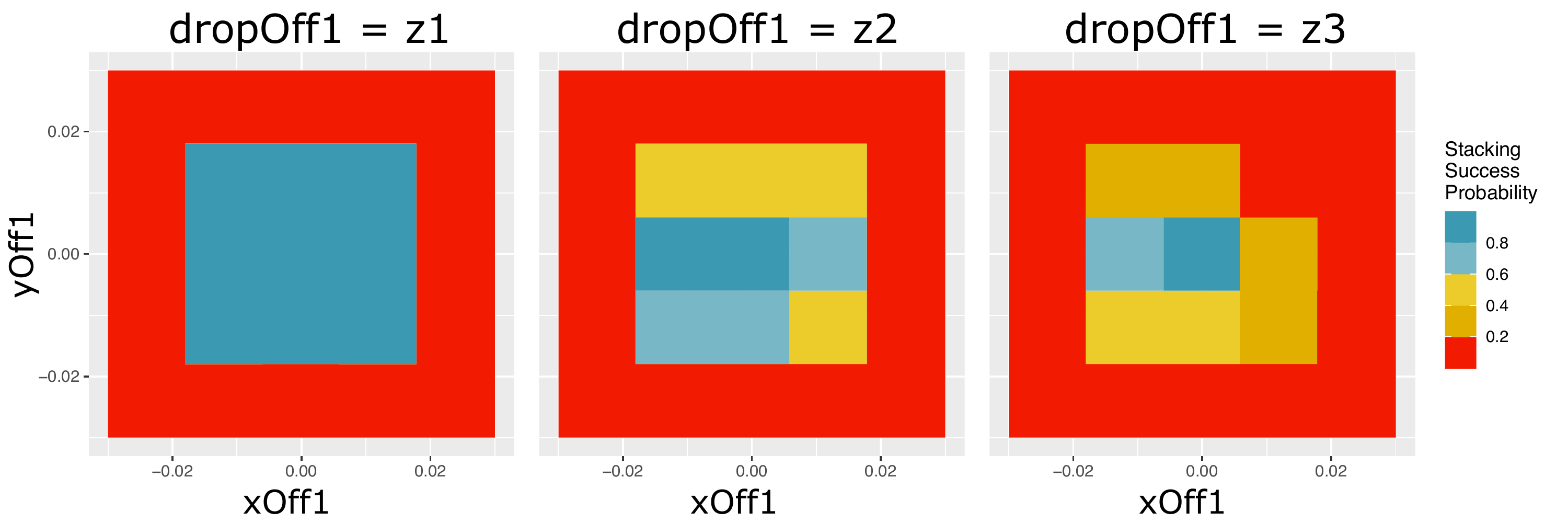}
  \caption{Visualisation of the conditional probability table for $P(\texttt{onTop}_1=1|\Pi_{\texttt{onTop}_1})$. $\texttt{xOff}_1$, $\texttt{yOff}_1$ are discretized into 5 intervals and $\texttt{dropOff}_1$.
  Values for $\texttt{xOff}_1$, $\texttt{yOff}_1$ are in meter.}
\label{fig:E1success}
\end{figure}

Fig.~\ref{fig:BNE1} visualizes the obtained causal relations between the subset of variables $\mathrm{\textbf{X}}_{E1} = \{ \texttt{xOff}_1$, $\texttt{yOff}_1$, $\texttt{dropOff}_1$ $\texttt{onTop}_1 \}$. The results exhibit dependencies of all the cube positioning variables ($\texttt{dropOff}_1, \texttt{xOff}_1, \texttt{yOff}_1$) on the stacking outcome $\texttt{onTop}_1$. Notice, that all the variables from $\mathrm{\textbf{X}}_{E1}$ were discretized according to Tab.~\ref{tab:intervalsE1}. For example, the variable of $\texttt{dropOff}_1$ has three possible intervals ($z_1, z_2, \text{and}\ z_3$). 
Fig.~\ref{fig:E1success} visualizes the obtained probabilities for the stacking success of $\textit{CubeUp}_1$ conditional on the analyzed variables. Generally, for all three drop-offset intervals, the causal model showed large stacking success probabilities centered around smaller x/y-offsets and decreases the larger the x/y-offsets become. This decrease in probability is faster for higher drop-offset positions (e.g., compare the two drop-offset intervals of $z_1$ and $z_3$). For the most extreme x/y-offset intervals, the model displays a stacking success of below 0.2 (red areas in Fig.~\ref{fig:E1success}), which is credible considering that the center of gravity of the stacked cubes in these x/y-offset intervals is close to the limits or outside the surface of the bottom cube.
We conclude that the probability distributions trained on simulated data are plausible.



\begin{table}[h]
\begin{tabular}{|l|l|l|}
\hline
$\texttt{dropOff}_1$ (in m) & $\texttt{xOff}_1$ (in m) & $\texttt{yOff}_1$ (in m) \\
\hline
\hline
$z_1:[0.005,0.037]$ & $x_1:[-0.03,-0.018]$ & $y_1:[-0.03,-0.018]$ \\
\hline
$z_2:(0.037,0.068]$ & $x_2:(-0.018,-0.006]$ & $y_2:(-0.018,-0.006]$ \\
\hline
$z_3:(0.068,0.100]$ & $x_3:(-0.006,0.0058]$ & $y_3:(-0.006,0.006]$ \\
\hline
 &   $x_4:(0.0058,0.0178]$  & $y_4:(0.006,0.018]$ \\
\hline
 &    $x_5:(0.0178,0.03]$  &  $y_5:(0.018,0.03]$ \\
\hline
\end{tabular}
\caption{Enlists the obtained discretization intervals for the variables $\mathbf{X}_{E1}$.}
\label{tab:intervalsE1}
\end{table}

\subsubsection{Learned causal model for Experiment 2}
\begin{figure}[ht!]
\centering
  \includegraphics[width=\textwidth]{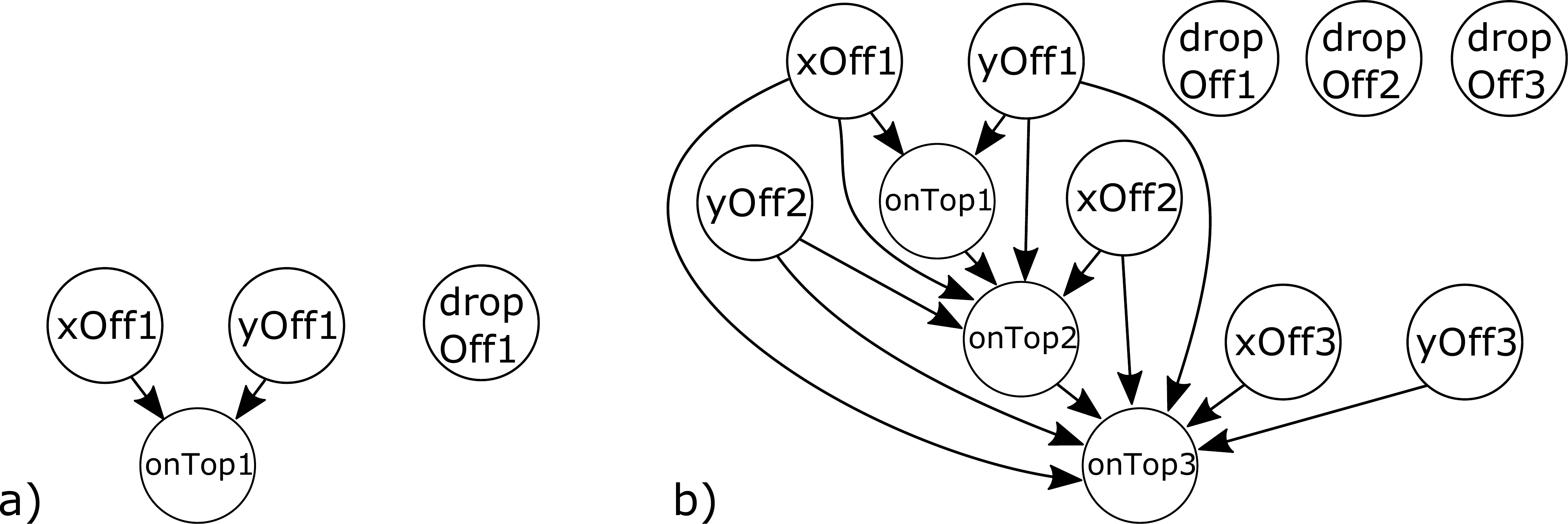}
  \caption{Obtained Bayesian network structure for the 3-Stack-Cube scenario. In a) the causal model obtained from only the first stacking action is compared with b) the complete causal model covering all three stacks.}
\label{fig:BNE2}
\end{figure}

 The obtained DAGs for the two evaluation cases ($\texttt{1-Stack-Model}_g$ and $\texttt{3-Stack-Model}_g$), are displayed in Fig.~\ref{fig:BNE2}. For both cases, the used variables were discretized according to Tab.~\ref{tab:intervalsE2}.
 In the case of the $\texttt{1-Stack-Model}_g$ ($i=1$), we obtained a slightly different dependency structure than in the model obtained from experiment 1 $\texttt{1-Stack-Model}_u$. In particular, we notice from Fig. \ref{fig:BNE2} that the drop-offset variable $\texttt{dropOff}_i$ is now independent from the stacking outcome $\texttt{onTop}_i$ compared to the obtained model shown in Fig. \ref{fig:BNE1} due to the different sampling distributions. In the new model, the Gaussian distribution for $\texttt{dropOff}_i$ samples created more values around the mean of 1cm, and smaller drop-offsets are shown to not have any measurable impact on the stacking success. 
 
 A similar graph dependency structure between the variables is observed in the $\texttt{3-Stack-Model}_g$ ($i={1,2,3}$). Where the $\texttt{dropOff}_i$ variables are independent of the stacking success. From Fig.~\ref{fig:BNE2}.b it becomes evident that the success of each stack depends on an increasing number of parent nodes. 
 Interestingly, the complete graph shown in Fig.~\ref{fig:BNE2}.b indicates that not only the x/y-offset variables but also previous $\texttt{onTop}_i$ impact the stacking success (e.g. consider the arrow from $\texttt{onTop}_2$ to $\texttt{onTop}_3$). The reason for these causal links is the termination of the stacking experiments in cases where the previous stack has already failed. We conclude that the $\texttt{3-Stack-Model}_g$ successfully captures the dependence of earlier stacks on the outcome of later stacking actions, thus encoding the action history in one causal model\footnote{Note that we do not visualize the conditional probability table for this experiment due to the increased number of variables that the stacking outcome is conditioned on. }. 
 
\begin{table}[h]
\begin{tabular}{|l|l|l|}
\hline
$\texttt{dropOff}_{1,2,3}$ (in m) & $\texttt{x/yOff}_{1,2}$ (in m) & $\texttt{x/yOff}_{3}$ (in m) \\
\hline
\hline
$z_1:[0.005,0.019]$ & $x_1:[-0.03,-0.014]$ & $y_1:[-0.03,-0.007]$ \\
\hline
$z_2:(0.019,0.036]$ & $x_2:(-0.014,-0.004]$ & $y_2:(-0.007,0.007]$ \\
\hline
$z_3:(0.036,0.1]$ & $x_3:(-0.004,0.004]$ & $y_3:(0.007,0.03]$ \\
\hline
 &   $x_4:(0.004,0.014]$  & \\
\hline
 &    $x_5:(0.014,0.03]$  &  \\
\hline
\end{tabular}
\caption{Enlists the obtained discretization intervals for the variables $\mathbf{X}_{E2}$.}
\label{tab:intervalsE2}
\end{table}

\subsection{Evaluation of Prediction and Failure Prevention Capabilities}
We finally analyze the ability of our obtained causal models to predict and avoid potential future failures by correcting the cubes' stacking position to the closest variable parametrization (as explained in Sec.~\ref{sec:correction}).
\subsubsection{Failure prediction and prevention in Experiment 1}
The results from the test datasets of experiment 1 are displayed in Tab.~\ref{tab:E1table}. Out of 40,000 collected samples of the ground-truth dataset, almost 30,000 stacking experiments failed (around $74\%$) without corrective actions. However, our model could fix and prevent $97\%$ of these failures. If we consider the confusion matrix in Tab.~\ref{tab:E1confusionMatrix}, we see that our model predicts almost all actual failures as failures (true negatives). At the same time, it predicts around 42\% ($11.1/26.2$) of the actually correct cases as failures (false negatives). For the current example, however, this might be beneficial since we want to avoid stacking failures and might not mind an additional action as much as a failed tower. We, therefore, conclude that our model is highly beneficial for predicting and avoiding future errors. 

\begin{table}[h]
\begin{tabular}{|l|l|l|}
\hline
Single Stack Case & failure percentage & corrected failures     \\
\hline
\hline
no model    & 74\% & 0\% \\
\hline
$\texttt{1-Stack-Model}_u$        &  1.9\% & 97\% \\
\hline
\end{tabular}
\caption{Percentage of failed stacking actions in Experiment 1 (40,000 data samples). Note that the $\texttt{1-Stack-Model}_u$ in this experiment is based on uniformly distributed data, as opposed to the $\texttt{1-Stack-Model}_g$ in experiment 2 (Table \ref{tab:E2results}), which is based on Gaussian distributions.}
\label{tab:E1table}
\end{table}

\begin{table}[h]
\begin{tabular}{|l|l|l|l|}
\hline
 & predicted as correct & predicted as failure &      \\
\hline
actually correct   & 15.1\% & 11.1\% & 26.2\%  \\
\hline
actually a failure  & 0.7\% & 73.1\% & 73.8\% \\
\hline
  & 15.8\% & 84.2\% & 100\% \\
\hline
\end{tabular}
\caption{Confusion Matrix for success prediction.}
\label{tab:E1confusionMatrix}
\end{table}

\subsubsection{Failure prediction and prevention in Experiment 2}
Failure prediction and correction ability are analyzed in Tab.~\ref{tab:E2results}. Without any corrective actions regarding the stacking positions, we sampled around $81\%$ stacking failures, of which about $21\%$ happened during the first stacking action, $42\%$ happened during the second, and $36\%$ during the third in the ground truth dataset (no corrections). This allows the conclusion that the failure probability increases with the height of the tower. The decreasing number of errors for the third stacking action is due to the limited number of data samples that reach the third stack. Considering the failure prediction confusion matrix of the $\texttt{3-Stack-Model}_g$ in Tab.~\ref{tab:E2confusionMatrix}, we can again detect a large true negative rate but also a high false negatives rate. If we could increase the number of discretization intervals, we could enhance the prediction capabilities of the model. However, an increased number of intervals also requires an increased number of training samples. As the second row of Tab.~\ref{tab:E2results} shows, applying the $\texttt{1-Stack-Model}_g$ on all three stacking actions reduced the number of stacking failures significantly by $78\%$ and applying the $\texttt{3-Stack-Model}_g$ reduces the errors by $75\%$. We can conclude that both models greatly improve the stacking success. Remarkably, the $\texttt{1-Stack-Model}_g$ performed well on all three stacking actions, despite having been trained only from data of the first stacking action and requiring 20 times fewer data than the $\texttt{3-Stack-Model}_g$. This is a good indication of the scalability possibilities of our model.

\begin{table}[h]
\begin{tabular}{|l|l|l|l|l|}
\hline
Three Stacks Case               &  failure percentage   & fail in 1 & fail in 2 & fail in 3 \\
\hline
\hline
no correction                   &  81\% & 21\%            & 42\%        & 36\%         \\
\hline
$\texttt{1-Stacks-Model}_g$ &         18\%     &              4\%     &           12\%          &            83\%         \\
\hline
$\texttt{3-Stacks-Model}_g$  & 20\% &  11.5\%            & 5.5\%             & 83\%   \\
\hline
\end{tabular}
\caption{Percentage of failed stacking actions in Experiment 2 (40,000 data samples). Note that the $\texttt{1-Stack-Model}_g$ in this experiment is based on Gaussian distributed data, as opposed to the $\texttt{1-Stack-Model}_u$ in experiment 1, which is based on uniformly distributed data.}
\label{tab:E2results}
\end{table}
\begin{table}[h]
\begin{tabular}{|l|l|l|l|}
\hline
 & predicted as correct & predicted as failure &      \\
\hline
actually correct   & 1.8\% & 17.3\% & 19.1\%  \\
\hline
actually a failure  & 0.3\% & 80.6\% & 80.9\% \\
\hline
  & 2.1\% & 97.9 \% & 100\% \\
\hline
\end{tabular}
\caption{Confusion Matrix for success prediction of $\texttt{3-Stacks-Model}_g$.}
\label{tab:E2confusionMatrix}
\end{table}

\subsection{Explanation \& Correction of timely shifted action effects}
\begin{figure}[ht!]
\centering
  \includegraphics[width=\textwidth]{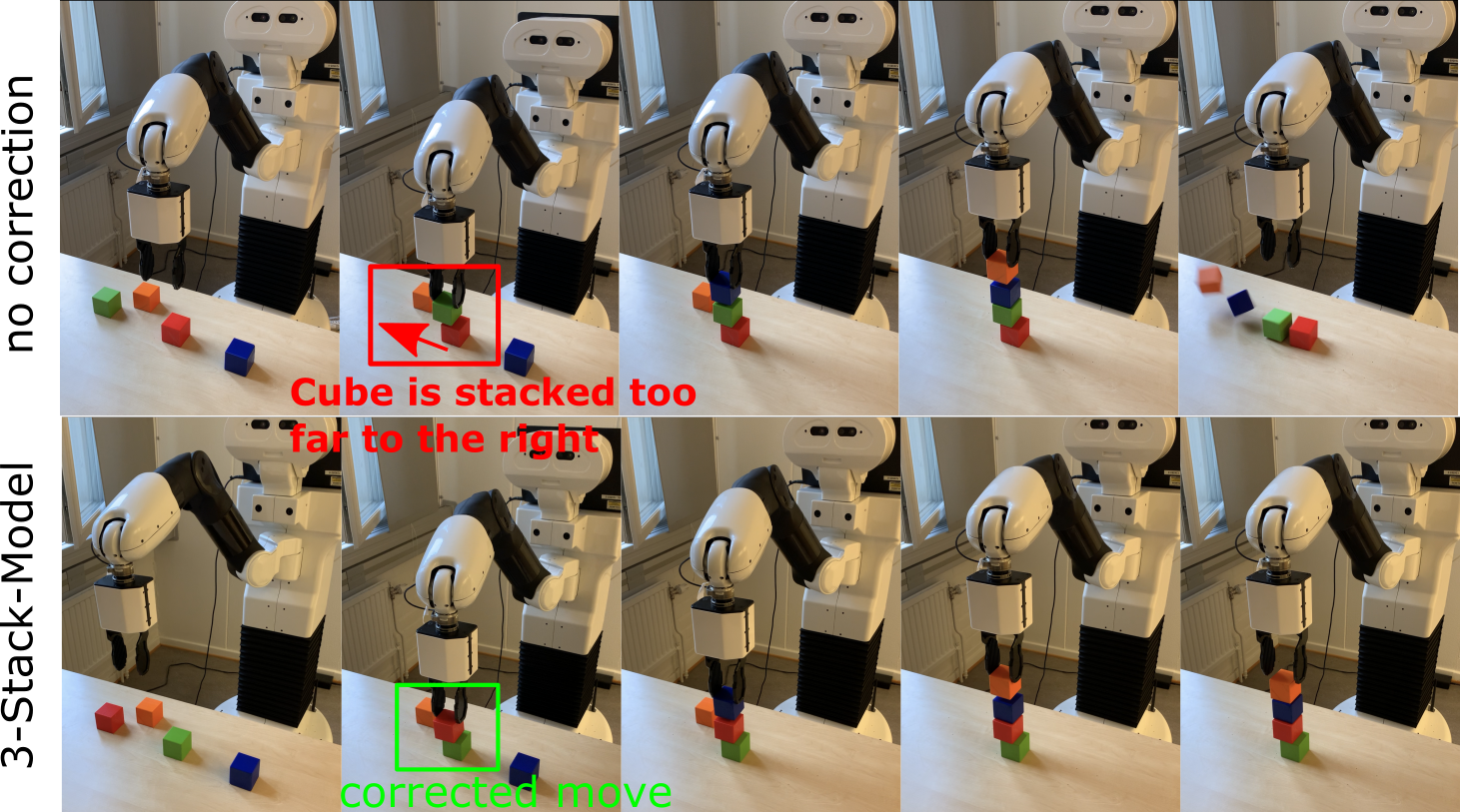}
  \caption{Displayes the stacking execution of Example 1 performed on the real robot. In the first row, the $\texttt{cubeUp}_1$ (green), is stacked too far to the right, which leads to a failure in the third stack. In the corrected sequence, as proposed by the $\texttt{3-Stack-Model}_g$, $\texttt{cubeUp}_1$ (red) is stacked a little more to the left, which allows the robot to successfully stack all 3 cubes.}
\label{fig:realStack}
\end{figure}

\begin{figure}[ht!]
\centering
  \includegraphics[width=\textwidth]{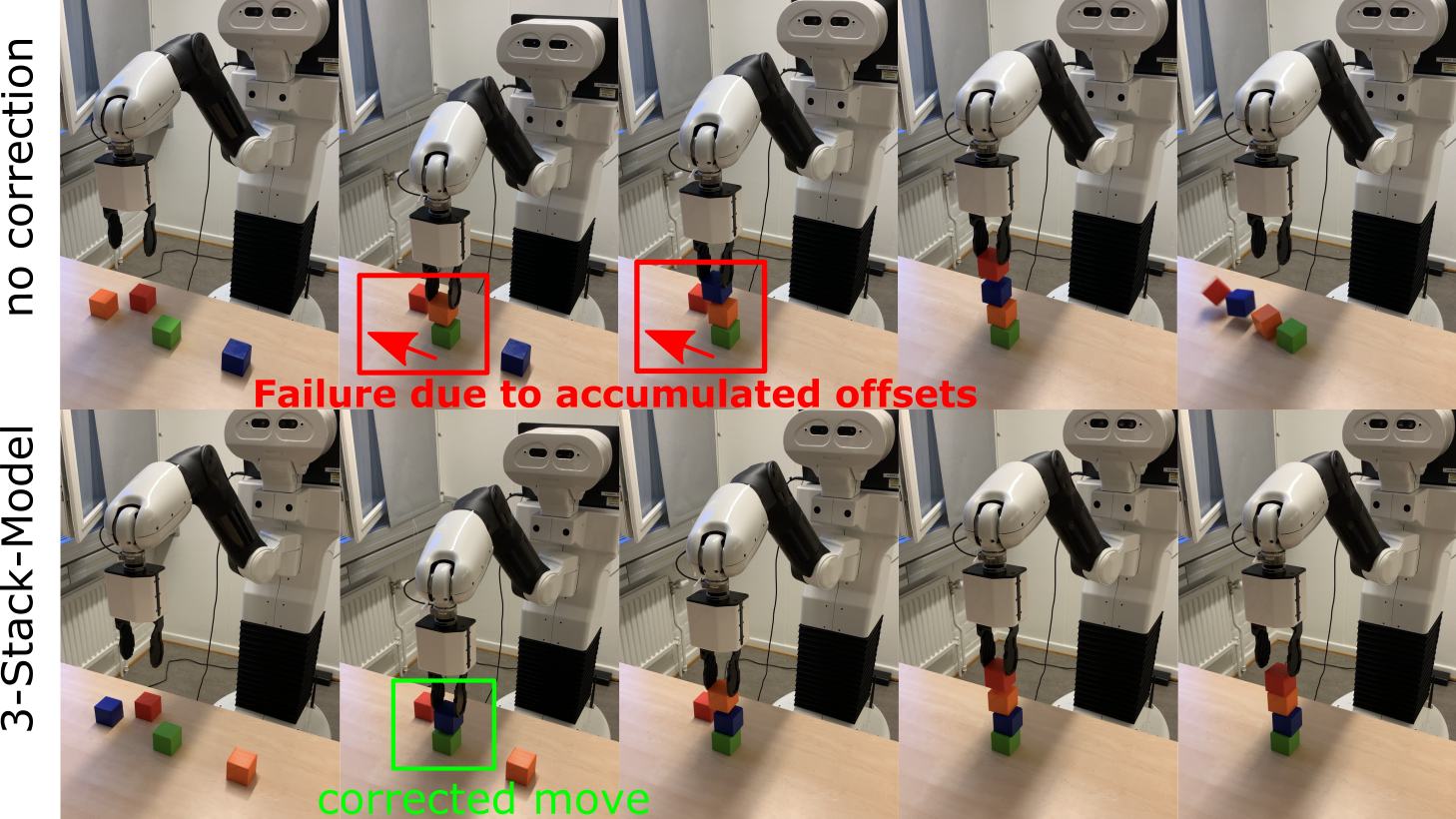}
  \caption{Displayes the stacking execution of Example 2 performed on the real robot. In the first row, each cube is stacked a little to the right and down. Each stack on itself is not found to be problematic by $\texttt{1-Stack-Model}_g$, and thus no corrective actions are found. However, with the third cube, the cumulative error is too large and the tower falls. In the corrected sequence, as proposed by the $\texttt{3-Stack-Model}_g$, $\texttt{cubeUp}_1$ (blue) is stacked a little more to the left, which allows the robot to successfully stack all 3 cubes, despite further offsets in the other stacks.} 
\label{fig:realStack2}
\end{figure}

Even though the $\texttt{3-Stack-Model}_g$ and the $\texttt{1-Stack-Model}_g$ perform approximately similar in terms of failure reduction, we required significantly more data samples to learn the structure and parameters of the BN. A critical advantage of the 3-Stack-Model, however, is the possibility to explain timely shifted action effects, as demonstrated through several examples in Tab.~\ref{tab:examples}. We set the probability threshold which distinguishes a failure from success to $\epsilon = 0.8$.
\begin{table}[h]
\begin{tabular}{|l|l|l|l|l|}
\hline
input & \begin{tabular}[c]{@{}l@{}}input\\interval\end{tabular} & \begin{tabular}[c]{@{}l@{}}curr. succ. \\ probability \end{tabular} & \begin{tabular}[c]{@{}l@{}} closest sol- \\ution interval \end{tabular}& \begin{tabular}[c]{@{}l@{}}exp. succ. \\ probability \end{tabular}\\
\hline
\hline
\multicolumn{5}{|c|}{\textbf{Example 1:}}\\
\hline
\begin{tabular}[c]{@{}r@{}}xOff1 = 0.02\\ yOff1 = 0.0\\ dropOff1 = 0.01 \\ xOff2 = 0.01\\ yOff2 = 0.0\\ dropOff2 = 0.01 \\ xOff3 = 0.0 \\ yOff3 = 0.0\\ dropOff3 = 0.01\end{tabular} & \begin{tabular}[c]{@{}l@{}}$\boldsymbol{x_5}$\\ $y_3$\\ $z_1$ \\ $x_3$\\ $y_3$\\ $z_1$ \\ $x_3$\\ $y_3$\\ $z_1$\end{tabular}  & 0.36 & 
\begin{tabular}[c]{@{}l@{}}$\boldsymbol{x_4}$\\ $y_4$\\ $z_1$ \\ $x_3$\\ $y_3$\\ $z_1$ \\ $x_3$\\ $y_3$\\ $z_1$\end{tabular} & 0.97 \\
\hline
\multicolumn{5}{|l|}{\begin{tabular}[c]{@{}l@{}}\textbf{Explanation}: The first cube was stacked too far to the right. \end{tabular}} \\
\hline
\hline
\multicolumn{5}{|c|}{\textbf{Example 2: }}\\
\hline
\begin{tabular}[c]{@{}l@{}}xOff1 = 0.01\\ yOff1 = -0.01\\ dropOff1 = 0.01 \\ xOff2 = 0.01\\ yOff2 = -0.01\\ dropOff2 = 0.01 \\ xOff3 = 0.01 \\ yOff3 = -0.01\\ dropOff3 = 0.01\end{tabular} & \begin{tabular}[c]{@{}l@{}}$\boldsymbol{x_4}$\\ $y_2$\\ $z_1$ \\ $x_4$\\ $y_2$\\ $z_1$ \\ $x_4$\\ $y_2$\\ $z_1$\end{tabular}  & 0.4 & \begin{tabular}[c]{@{}l@{}}$\boldsymbol{x_3}$\\ $y_2$\\ $z_1$ \\ $x_4$\\ $y_2$\\ $z_1$ \\ $x_4$\\ $y_2$\\ $z_1$\end{tabular} & 0.88 \\
\hline
\multicolumn{5}{|l|}{\begin{tabular}[c]{@{}l@{}}\textbf{Explanation}: The first cube was stacked too far to the right. \end{tabular}} \\
\hline
\end{tabular}
\caption{Two examples for failure prevention. Corresponding real-world experiments are visualized in Fig. \ref{fig:realStack} and \ref{fig:realStack2}.}
\label{tab:examples}
\end{table}
We confirmed each of the examples by performing six real-world experiments per example; three for the initial variable parametrization and three for the correction proposed by the $\texttt{3-Stack-Model}_g$. 
The examples showcase scenarios where the tower did not fall directly when the error had been committed but only became evident later. In example 1 (see Tab.~\ref{tab:examples}, Fig.~\ref{fig:realStack}), the first cube has been stacked far to the left. This did not lead to a failure immediately, but after stacking the second (1 out of 3 times) or third cube (2 out of 3 times), the tower fell. Our model detected that the culprit was this first stack and not the third. Taking each stack individually would have also yielded a correction of the first stacking action since the $\texttt{1-Stack-Model}_u$\footnote{In reality, the underlying distribution of variables like $\texttt{xOff}_i$ might be unknown. Therefore we cannot know if the model trained on a uniform or Gaussian data is more appropriate.} predicts a success probability of $< 0.2$ (Fig. \ref{fig:E1success}), despite the tower not failing in reality. As discussed in Tab.\ref{tab:E1confusionMatrix}, the model has a relatively large number of false negatives, which, however, in this case, was beneficial to unintentionally prevent a future failure. However, in example 2 (see Tab.~\ref{tab:examples}, Fig.~\ref{fig:realStack2}), we investigated a case where the $\texttt{1-Stack-Model}_u$ did not predict any errors since each stack looks perfectly fine on its own. Only due to the cumulative effect of several offsets did we get a failure at the last stack of cube 3 (3 out of 3 cases). Without the history of the 3-Stack-Model, we would not be able to correct this example. In both examples, the corrected version by the $\texttt{3-Stack-Model}_g$ succeeded in all three trials\footnote{\textbf{The robot executions can be seen in \url{https://youtu.be/baM2hw4piv8}}}. 


\section{Conclusion}
\label{sec:conclusion}
In this paper, we propose a causal-based method that allows robots to understand possible causes for execution errors and predict how likely an action will succeed. We then introduce a novel method that utilizes these prediction capabilities to find corrective actions which will allow the robot to prevent failures from happening. Our algorithm proposes a solution to the complex challenge of timely shifted action effects. By detecting causal links over the history of several actions, the robot can effectively predict and prevent failures even if the root of a failure lies in a previous action. We have shown the success of our approach for the problem of stacking cubes in two cases; a) single stacks (stacking one cube) and; b) multiple stacks (stacking three cubes). In the single-stack case, our method was able to reduce the error rate by $97\%$. We also show that our approach can scale to capture multiple actions in one model, allowing to measure timely shifted action effects, such as the impact of an imprecise stack of the first cube on the stacking success of the third cube. For these complex situations, our model was able to prevent around $75\%$. 

Despite being able to capture action histories in one model, one disadvantage of such large models is data efficiency. The more parents a BN node has, the more samples are required to learn its graphical structure and the conditional probabilities. In the future, one important aspect of the feasibility of such models is to find intelligent ways of pre-initializing the model, e.g., by utilizing the single-action models as prior for structure and probabilities.




 \bibliographystyle{elsarticle-num} 
 \bibliography{cas-refs}





\end{document}